\documentclass[letterpaper, 10 pt, journal, twoside]{IEEEtran}
\usepackage{amsmath}
\usepackage{algorithmic}
\usepackage{array}
\usepackage{textcomp}
\usepackage{stfloats}
\usepackage{url}
\usepackage{verbatim}
\usepackage{graphicx}  % for rotatebox
\usepackage{multirow}  % for multirow cells
\usepackage{color, array}
\definecolor{MyColor}{gray}{0.75}
\usepackage{colortbl}
\usepackage{balance}                                 % Needed to meet printer requirements.

\usepackage{graphics} % for pdf, bitmapped graphics files
\usepackage{epsfig} % for postscript graphics files
\usepackage{xcolor}
\usepackage{amssymb}
\usepackage{adjustbox}
\usepackage{booktabs} % To thicken table lines
\usepackage[ruled,vlined]{algorithm2e}
\usepackage{hyperref}
\usepackage{xcolor}
\usepackage{graphicx}
\usepackage{caption}
\usepackage{amsmath,amssymb, fancyhdr, color, comment, graphicx, environ}
\usepackage{xcolor}
\usepackage{mdframed}
\usepackage{indentfirst}
\usepackage{hyperref}
\usepackage{subcaption}
\usepackage{hyperref}
\usepackage{wrapfig}
\usepackage{ltablex}% <-- added
\usepackage{siunitx}% <-- added
\usepackage{caption}% <-- added
\newcolumntype{L}{>{\raggedright\arraybackslash}X}% <-- added
\usepackage{svg}

\newcommand{\ie}{\textit{i.e.}}
\newcommand{\eg}{\textit{e.g.}}
\newcommand{\wrt}{\textit{w.r.t.\ }}

\def  \leftvertical {0.0} % [cm] crop 
\def  \rightvertical {2.5} % [cm] crop 
\def  \bottom {0.5} % [cm] crop the bottom 
\def  \top {2.5} % [cm] crop the top 

\begin{document}

\title{PointNetPGAP-SLC: A 3D LiDAR-based Place Recognition Approach with Segment-level Consistency Training for Mobile Robots in Horticulture}

\author{T. Barros$^{1}$, L. Garrote$^{1}$, P. Conde$^{1}$, M.J. Coombes$^{2}$, C. Liu$^{2}$, C. Premebida$^{1}$, U.J. Nunes$^{1}$%
\thanks{$^{1}$T.Barros, L.Garrote, P.Conde, C.Premebida and U.J.Nunes are with the University of Coimbra, Institute of Systems and Robotics, Department of Electrical and Computer Engineering, Portugal.
E-mails:{\tt\small\{tiagobarros,~garrote,~pedro.conde,
~cpremebida,~urbano\}@isr.uc.pt}}%
\thanks{$^{2}$M.J.Coombes and C.Liu are with the Dept. of Aeronautical and Automotive Engineering, LUCAS Lab, Loughborough University, UK.
E-mails:{\tt\small\{M.J.Coombes,~C.Liu5\}@lboro.ac.uk}}%
}

%\markboth{IEEE Robotics and Automation Letters. Preprint Version.}
\markboth{Preprint Version.}
{Barros \MakeLowercase{\textit{et al.}}: A 3D LiDAR-based Place Recognition Approach  with SLC Training for Mobile Robots in Horticulture}

\twocolumn[{%
\renewcommand\twocolumn[1][]{#1}%
\maketitle
%\vspace{-1cm}
\begin{center}
    \centering
    \captionsetup{type=figure}
    \includegraphics[width=\textwidth]{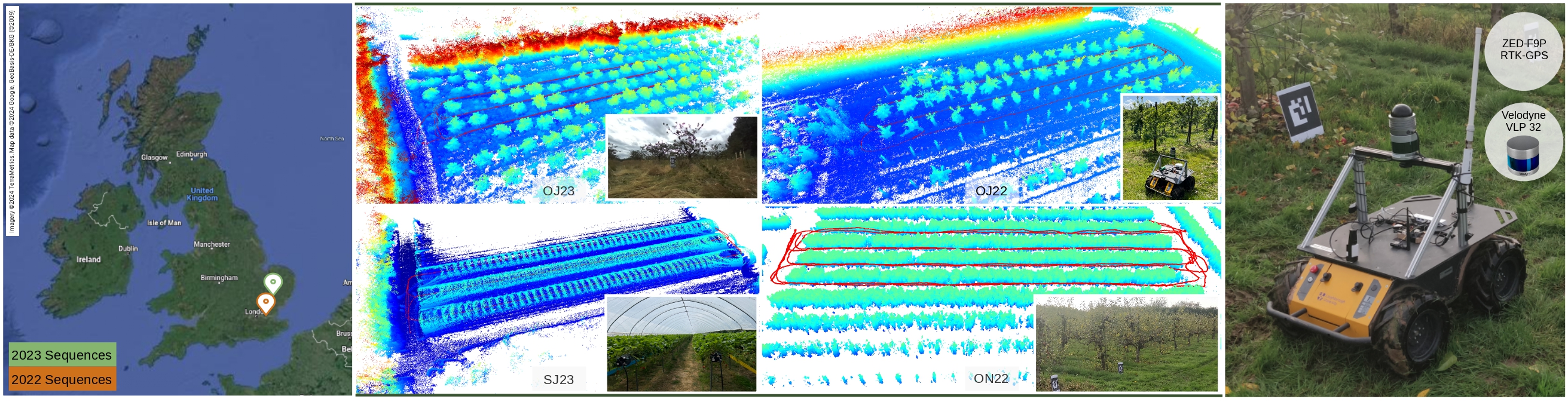}
    \captionof{figure}{Geolocation, 3D maps and recording setup of the HORTO-3DLM dataset. Four data sequences were recorded over two years at different locations in the UK, using a Husky mobile platform equipped with a 32-beam Velodyne sensor and ZED-F9P RTK GPS/GNSS system.} %
    \label{fig:fronte}
\end{center}%
}]

\renewcommand{\thefootnote}{\fnsymbol{footnote}}
%\footnotetext[0]{Manuscript received: May, 24, 2024; Revised: August, 20, 2024; Accepted: September, 18, 1024.} 
%\footnotetext[0]{This paper was recommended for publication by Editor Sven Behnke upon evaluation of the Associate Editor and Reviewers’ comments.} 
\footnotetext[0]{This work has been supported by the project GreenBotics (ref. PTDC/EEI-ROB/2459/2021), funded by Fundação para a Ciência e a Tecnologia (FCT), Portugal, and by Innovate UK under the Grant No. 10072930 and 10073332. It was also partially supported by FCT through grant UIDB/00048/2020 and under the PhD grant with reference 2021.06492.BD.}
\footnotetext[0]{\vspace{0cm}$^1$T.Barros, L.Garrote, P.Conde, C.Premebida and U.J.Nunes are with the University of Coimbra, Institute of Systems and Robotics, Department of Electrical and Computer Engineering, Portugal.
E-mails: {\tt\footnotesize \emph\{tiagobarros,~garrote,~pedro.conde,
~cpremebida,~urbano\}@isr.uc.pt}}
\footnotetext[0]{\vspace{0cm}$^2$M.J.Coombes and C.Liu are with the Dept. of Aeronautical and Automotive Engineering, LUCAS Lab, Loughborough University, UK.
E-mails:{\tt\small\{M.J.Coombes,~C.Liu5\}@lboro.ac.uk}}
%\footnotetext[0]{Digital Object Identifier (DOI): see top of this page.}% %
%%%%%%%%%%%%%%%%%%%%%%%%%%%%%%%%%%%%%%%%%%%%%%%%%%%%%%%%%%%%%%%%%%%%%%%%%%%%%%%% 

\begin{abstract}
3D LiDAR-based place recognition remains largely underexplored in horticultural environments, which present unique challenges due to their semi-permeable nature to laser beams. This characteristic often results in highly similar LiDAR scans from adjacent rows, leading to descriptor ambiguity and, consequently, compromised retrieval performance.
In this work, we address the challenges of 3D LiDAR place recognition in horticultural environments, particularly focusing on inter-row ambiguity by introducing three key contributions: (i) a novel model, PointNetPGAP, which combines the outputs of two statistically-inspired aggregators into a single descriptor; (ii) a Segment-Level Consistency (SLC) model, used exclusively during training to enhance descriptor robustness; and (iii) the HORTO-3DLM dataset, comprising LiDAR sequences from orchards and strawberry fields.
Experimental evaluations conducted on the HORTO-3DLM and KITTI Odometry datasets demonstrate that PointNetPGAP outperforms state-of-the-art models, including OverlapTransformer and PointNetVLAD, particularly when the SLC model is applied. These results underscore the model's superiority, especially in horticultural environments, by significantly improving retrieval performance in segments with higher ambiguity. The dataset and the code will be made publicly available at \href{https://github.com/Cybonic/PointNetPGAP-SLC.git}{https://github.com/Cybonic/PointNetPGAP-SLC.git}
\end{abstract}

\begin{IEEEkeywords}
Deep Learning Methods; Localization; Agricultural Automation
\end{IEEEkeywords}

\IEEEpeerreviewmaketitle

\section{INTRODUCTION}
\IEEEPARstart{I}{n} recent years, significant advancements have been made in 3D LiDAR-based place recognition, driven primarily by the autonomous driving community with the aim of achieving long-term localization~\cite{barros2021place,barros2023trer}. Although initial efforts have been made to adapt these methods for agricultural environments~\cite{guo2019local, cheng2023treescope}, several challenges unique to these settings remain largely unresolved. Unlike urban environments, where LiDAR scans capture well-defined geometries (\eg, corners, planes), horticultural environments produce scans with sparse and poorly defined features. The semi-permeable nature of these environments allows laser beams to traverse multiple row layers, resulting in scans that cover several rows. This phenomenon leads to sparse scans that overlap with scans from adjacent rows, thereby causing both intra- and inter-row descriptor ambiguity.

This work addresses the challenges of 3D LiDAR place recognition in horticultural environments, particularly focusing on inter-row ambiguity, by introducing three key contributions:
(i) PointNetPGAP: A lightweight model that combines PointNet for local feature extraction and integrates the outputs of two statistically-inspired aggregation approaches into a single descriptor.
(ii) Segment-Level Consistency (SLC) Model: A training model designed to reduce inter-row ambiguity by differentiating scans from different segments.
(iii) HORTO-3DLM Dataset: A novel dataset collected using a 3D LiDAR sensor, comprising data from three orchards and a strawberry plantation.

We benchmark PointNetPGAP at both the sequence and segment levels against state-of-the-art (SOTA) models, specifically OverlapTransformer~\cite{9785497}, PointNetVLAD~\cite{angelina2018pointnetvlad}, and LOGG3D-Net~\cite{9811753}, using the HORTO-3DLM dataset within a cross-validation protocol. Additionally, we evaluate the cross-domain generalization capabilities using the KITTI Odometry dataset~\cite{Geiger2012CVPR}.
The results highlight that the aggregation approach of PointNetPGAP is better suited to horticultural environments compared to existing SOTA models. Although its impact is reduced in urban settings, it nonetheless exhibits superior rotation invariance in urban environments compared to SOTA models. Moreover, when models are trained with the SLC approach, results at both the sequence and segment levels indicate improved retrieval performance, particularly in segments with higher ambiguity.

\begin{figure*}[pt]
    \centering
    \includegraphics[width=\textwidth,trim=0 0cm 0cm 0, clip]{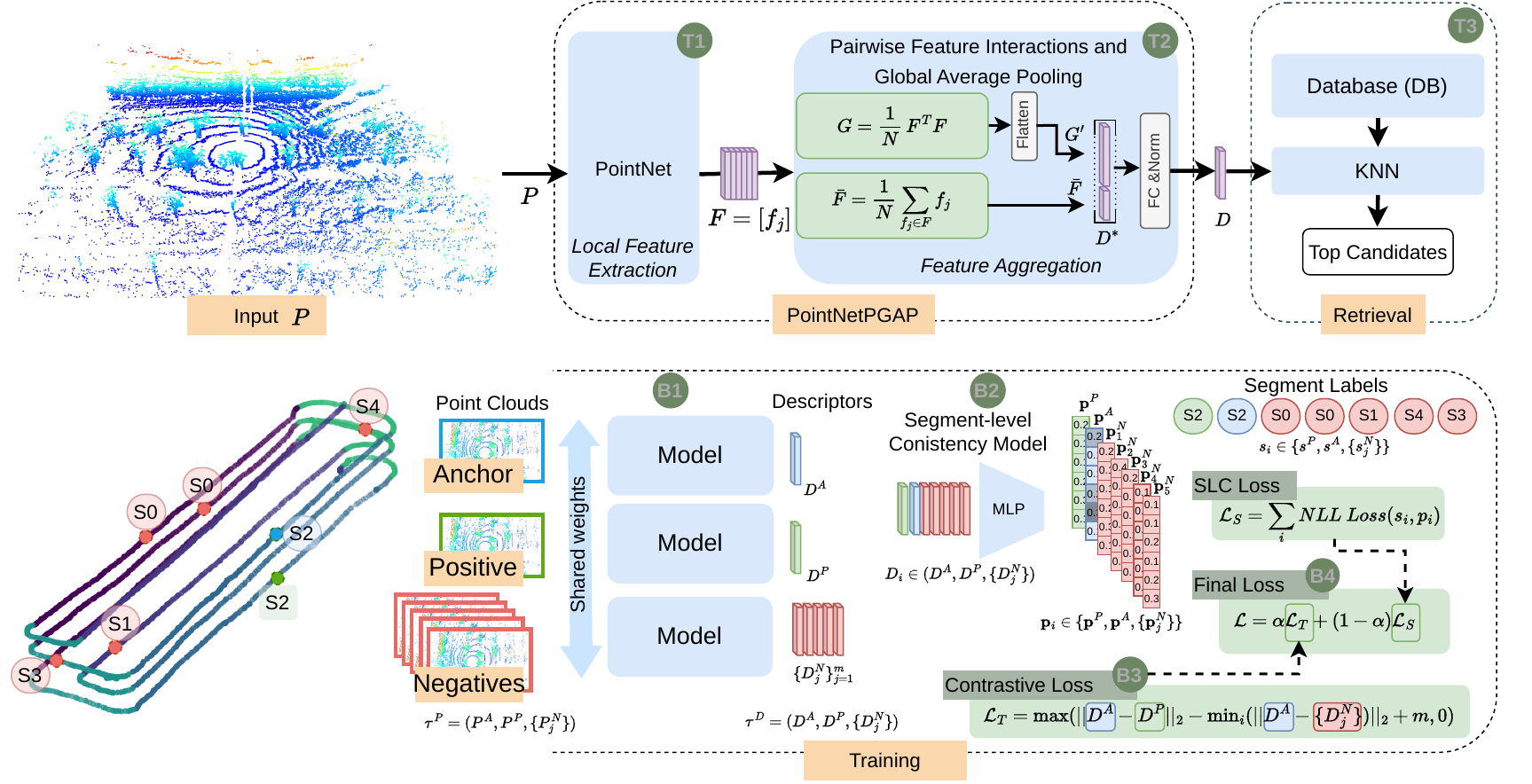}
    \caption{ \textbf{Top}: The retrieval framework uses PointNetPGAP to generate global descriptors through the following stages: T1) PointNet processes an input scan $P$ to extract local features $F$; T2) These features are aggregated into a global descriptor $D$ by combining global average representation with pairwise feature interactions; T3) The global descriptor is then used to query the database and retrieve the top-$k$ most similar places. \textbf{Bottom}: The training scheme involves: B1) Inputting a training tuple with an anchor-positive pair and $m$ negatives to generate descriptors; B2) Feeding these descriptors into an MLP in the SLC model to predict segment classes, the SLC loss computed via Negative Log-likelihood using the predictions and  the segment labels; B3) Calculating the LazyTriplet loss based on anchor-positive and anchor-negative distances; B4) Combining both losses with a weight  $\alpha$. }

    \label{fig:pipeline}
\end{figure*}

%%%========= New Section =========
\section{RELATED WORK}\label{sec:rw}
Place recognition has been the focus of extensive research in recent years, particularly within the autonomous driving community \cite{barros2021place}, where Deep Learning-based approaches have garnered the most attention. One potential reason for the limited research in other domains may be the scarcity of publicly available datasets.
\subsection{3D LiDAR-based Place Recognition}

In the context of 3D LiDAR-based methods, although hand-crafted descriptors are still being used~\cite{ou2023place}, Deep Learning (DL) techniques have emerged as the most prevalent choice for place modeling~\cite{barros2021place}. DL-based place recognition models can be broadly split into two main components: a local feature extraction module that maps the input scan to a local feature space, and a feature aggregation module that aggregates the local features into a global descriptor.

As for the feature extraction module, approaches can be categorized based on how the point clouds are processed: those that extract features directly from point clouds, such as PointNetVLAD~\cite{angelina2018pointnetvlad}, and those that resort to proxy representations, such as voxels~\cite{siva2020voxel,komorowski2021minkloc3d}, polar coordinates~\cite{li2022rinet}, or depth range images~\cite{barros2022attdlnet,9785497}. Recent works such as MinkLoc3D~\cite{komorowski2021minkloc3d}, LCDNet~\cite{cattaneo2022lcdnet}, and LOGG3D-Net~\cite{9811753} tend to primarily use voxel-based approaches, employing sparse convolutions such as PV-RCNN~\cite{shi2020pv} or 4D Spatio-Temporal ConvNets~\cite{choy20194d}, which have shown higher feature extraction capabilities in urban-like environments. In this work, we resort to PointNet~\cite{qi2017pointnet} due to its efficiency and lower computational demands (as demonstrated in Section~\ref{sec:runtime}).

Regarding aggregation strategies, most 3D LiDAR place recognition frameworks rely on first-order aggregation approaches such as NetVLAD~\cite{angelina2018pointnetvlad,arandjelovic2016netvlad,cattaneo2022lcdnet}, which computes the residuals \wrt feature clusters, generalized-mean pooling (GeM)~\cite{komorowski2021minkloc3d} or global max pooling (MAC)~\cite{tolias2015particular} which operate directly on the feature space. Inspired by these works, we propose an aggregation approach that combines, in a single descriptor, a first-order and a second-order aggregation approach. Specifically, we combine the global average pooling aggregator with a pairwise feature interaction aggregator into a single descriptor. We compare our hypothesis with the SOTA models LOGG3D-Net~\cite{9811753}, PointNetVLAD~\cite{angelina2018pointnetvlad}, and OverlapTransformer~\cite{9785497}, and show through empirical results on real-world datasets that the combination of these two statistical-based aggregators yields higher performance than each aggregator individually and also the SOTA models.

\subsection{Datasets}
3D LiDAR-based place recognition methods have gained traction as robust and scalable global localization approaches, primarily driven by the autonomous vehicle community, which works mainly on road/city environments~\cite{barros2021place}. Works such as MinkLoc3D~\cite{komorowski2021minkloc3d}, PointNetVLAD~\cite{angelina2018pointnetvlad}, LCDNet~\cite{cattaneo2022lcdnet}, or LOGG3D-Net~\cite{9811753} are evaluated on urban environments, using available 3D LiDAR urban datasets such as KITTI odometry~\cite{Geiger2012CVPR}, MulRan~\cite{kim2020mulran}, Oxford RobotCar~\cite{maddern20171,angelina2018pointnetvlad}, or KITTI-360~\cite{xie2016semantic}.

Less studied is 3D LiDAR place recognition in field or agricultural environments. One possible reason for this is the lack of available datasets which, recently, has been partially compensated by datasets such as Wild-places~\cite{knights2023wild}, RELLIS-3D~\cite{jiang2021rellis}, ORFD~\cite{min2022orfd}, TreeScope~\cite{cheng2023treescope}, or VineLiDAR~\cite{prabhu2024uavs}, showing the increasing interest in such environments. 

\section{Proposed Approach} \label{sec:pa}
This section introduces the PointNetPGAP and the proposed training regime, where a traditional contrastive learning approach is combined with the proposed SLC model. Figure~\ref{fig:pipeline} outlines both PointNetPGAP in a retrieval framework (\textbf{Top}) and PointNetPGAP in the proposed training scheme (\textbf{Bottom}). 

\subsection{Problem Formulation}
A retrieval-based  place recognition framework comprises two stages (as illustrated in Fig.~\ref{fig:pipeline}): firstly (denoted by T1 and T2), a LiDAR scan  $P \in \mathbb{R}^{n\times 3}$ with $n$ points is mapped to a descriptor $D \in \mathbb{R}^{d}$, such that $\Theta:\mathbb{R}^{n\times 3} \rightarrow \mathbb{R}^{d}$; then (the 2$^{nd}$ stage: T3), the descriptor $D$ queries a database for the $k$ nearest neighbors (KNN) based on a similarity metric.  The $k$ retrieved candidates represent the top-$k$ candidates for potential revisited places. In this work, we focus on the first part (\ie, T1 and T2) and follow a retrieval framework (\ie, T3) as described in~\cite{barros2022attdlnet}.

\subsection{PointNetPGAP} \label{sec:pointnetgap}
PointNetPGAP, defined here as a mapping function $\Theta$ (outlined in Fig.~\ref{fig:pipeline} by T1, T2), can be decomposed into two functions: $\Theta \equiv  \varphi \circ \phi$, where $\phi:\mathbb{R}^{n\times 3} \rightarrow  \mathbb{R}^{n \times c}$ maps  the input scan $P \in \mathbb{R}^{n\times 3}$ to hyper-dimensional features  $F \in \mathbb{R}^{n\times c}$ with $c$ dimensions; and $\varphi: \mathbb{R}^{n\times c} \rightarrow \mathbb{R}^{d}$ aggregates the local features $F$ into a global descriptor $D \in \mathbb{R}^{d}$ with $d$ dimensions. 

\subsubsection{Local Feature Extraction (T1)} In this work, we follow the same approach as in \cite{angelina2018pointnetvlad} and implement $\phi$ based on PointNet \cite{Qi_2017_CVPR}, a DL-based network that operates directly on LiDAR scans. Hence, it receives a LiDAR scan $P \in \mathbb{R}^{n\times 3}$ with $n$ points and returns a local feature representation $F \in \mathbb{R}^{n\times c}$ with $c$ dimensions.

\subsubsection{Feature Aggregation (T2)} As for the aggregation (\ie, $\varphi$), this work proposes the fusion of two statistically-inspired aggregators: a global average representation of the local features, also known as global average pooling, and an aggregator based on pairwise feature interactions. 

The  global average representation, hereafter referred to as GAP, is computed as follows:
\begin{equation}
    \bar{F} = \frac{1}{n} \sum_{f_j \in F} f_{j}, 
\end{equation}
\noindent  where $f_j \in \mathbb{R}^{c}$ represents the $j$-th feature vector, and  $\bar{F} \in \mathbb{R}^{c}$ represents the global average representation. As for the aggregator based on the pairwise feature interactions, it is computed as follows:
\begin{equation}
    G = \frac{1}{n} F^T\,F, 
\end{equation}
\noindent where $G \in \mathbb{R}^{c \times c}$ represents pairwise feature interactions of the local features, which is flattened into a vector representation $G' \in \mathbb{R}^{c^2}$. Finally, to create the final descriptor $D$, both representations are concatenated into a single vector $D^* = [G',\bar{F}] \in \mathbb{R}^{c+c^2}$, which is fed to a full-connected layer $\text{FC}: \mathbb{R}^{c+c^2} \rightarrow  \mathbb{R}^d$, and normalized using the $L_2$-norm: $D = L_2\text{-norm}(\text{FC}(D^*))$.

\subsection{Network Training}
Training PointNetPGAP involves a traditional contrastive approach using the LazyTriplet loss \cite{angelina2018pointnetvlad}, along with the proposed SLC model. The training process is illustrated at the bottom of Fig.~\ref{fig:pipeline} and is divided into four main stages, labeled B1, B2, B3, and B4.

\subsubsection{Training Data (B1)} \label{sec:b1} 
Place recognition methods traditionally rely on contrastive learning techniques, which are designed to differentiate between similar (positive pairs) and dissimilar (negative pairs) data samples. In urban environments, positive pairs are typically defined by a fixed radius, which is a sufficient condition given the scale of the trajectories and revisits in such environments. However, in narrow-row crop environments, a fixed radius is a necessary but not sufficient condition for place recognition. Additional constraints are required to ensure that loops occur within the same row, rather than between neighboring rows. To address this, crops are divided into distinct segments, and loops must not only satisfy the radius condition but also originate from the same segment.

\begin{figure*}[!t]
    \centering
    \includegraphics[width=1\textwidth, trim={0cm 0cm 0cm 0cm},clip]{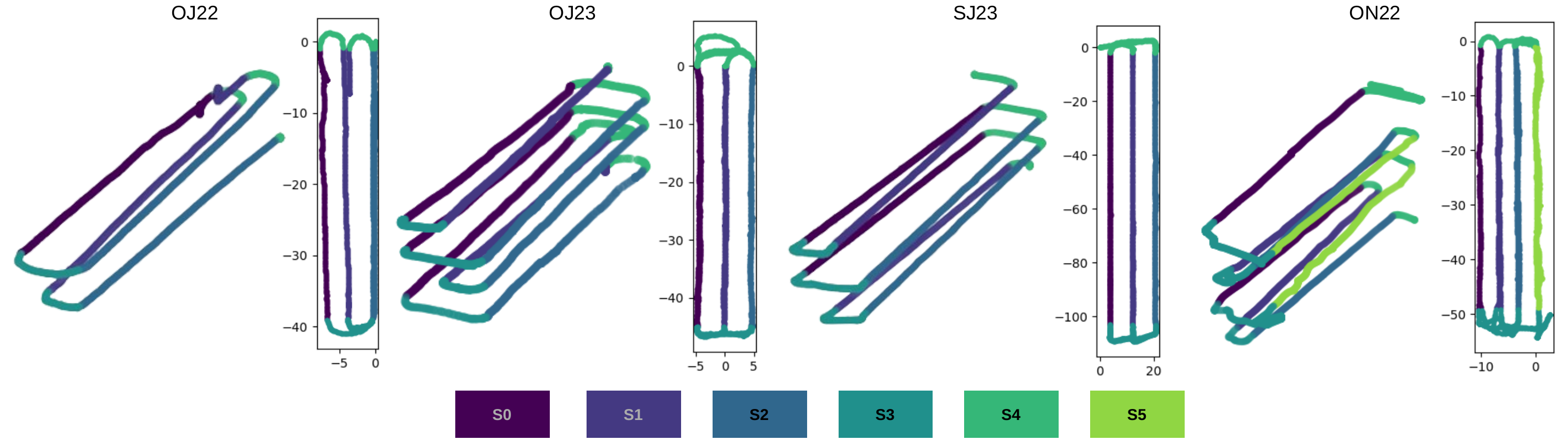}
    \caption{2D and 3D visualizations of the sequences' paths. The 2D representation shows the individual segments, while the 3D representation outlines the overlapping paths.}
    \label{fig:rows}
\end{figure*}

Hence, in this work, a positive pair refers to a pair of scans that must meet the following constraints: (1) both scans must be within a $r_{th}$ range (in meters); (2) both scans must be from different revisits, meaning that the positive scan cannot be the immediate previous scan of the anchor; and (3) both scans must be from the same segment (see Fig.~\ref{fig:rows}). Scan pairs that do not meet these three constraints belong to the set of negatives.

As such, the model is trained with tuples of scans (that satisfy the aforementioned conditions)  $\tau^P = (P^A,P^P,\{P^N_{j}\})$ where $P^A \in \mathbb{R}^{n \times 3}$ is the anchor, $P^P \in \mathbb{R}^{n \times 3}$ is the closest positive scan \wrt $P^A$ in the Euclidean space, while $\{P^N_{j}| P^N_{j} \in \mathbb{R}^{n \times 3}\}_{j=1}^m$ is a set of $m$ randomly selected negatives. The scan tuple $\tau^P$ is mapped to the descriptor space (as described in Section \ref{sec:pointnetgap}), generating the corresponding descriptor tuple $\tau^D = (D^A,D^P,\{D^N_j\})$.  

\subsubsection{Segment-level consistency model (B2)} 
The SLC model introduces an additional signal during training, which is more context-related than the one generated by the contrastive loss. The goal is to resolve ambiguities created by scans that are highly overlapped but belong to neighboring segments. The SLC model is framed as a classification problem, aiming to predict the segment class using the descriptors as inputs. The loss between the predictions and the corresponding ground-truth segment class $s \in \{1,2,...,L\}$, where $L$ is the number of segments in a sequence (as shown in Fig.~\ref{fig:rows}), is computed based on the NLL Loss. Hence, given a training tuple of descriptors  $\tau^D$, and the corresponding tuple of ground-truth segment classes $\tau^S = (s^A,s^P,\{s^N_j\})$,  all descriptors $D_i \in \tau^D$ are fed to an MLP to predict the corresponding segment class, such that:
\begin{equation}
    \mathbf{p}_i = \text{MLP}(D_i),
\end{equation}
\noindent where $\mathbf{p}_i = [p_1, p_2, ..., p_L]^T$ with $\sum\limits_{p_k \in \mathbf{p}_i} p_k = 1$ representing the segment prediction likelihood. The predictions are used, together with the respective ground-truth segment label $s_i \in \tau^S$, to compute the loss as follows:
\begin{equation}
    \mathcal{L}_S = \sum_i NLLLoss(\mathbf{p}_i,s_i).  
\end{equation}

\subsubsection{Contrastive Loss (B3)} 
As for the contrastive cost function, the LazyTriplet loss \cite{angelina2018pointnetvlad} is used. Thus, given a descriptor tuple $\tau^D = (D^A,D^P,\{D^N_j\})$ the loss is computed as follows:
\begin{equation}
    \mathcal{L}_T= \text{max}(d_{AP}-d_{AN}+m,0),
    \label{eq:lt}
\end{equation}

\noindent where $d_{AP} = \|D^A-D^P\|_2$ is the Euclidean distance between the anchor and the positive, $d_{AN} = \min\limits_{j} (\|D^A-\{D^N_{j}\}\|_2)$ is the Euclidean distance between the anchor and the hardest negative: \ie,  the negative that is the closest \wrt $D^A$ in the descriptor space, while $m$ designates a margin value.

\subsubsection{Final Loss (B4)} Finally, both $\mathcal{L}_T $ and $\mathcal{L}_S$ are combined as follows: 
\begin{equation}
     \mathcal{L} = \alpha \mathcal{L}_T + (1-\alpha) \mathcal{L}_S,
\end{equation}
\noindent where  $\alpha$ is adjustable weight.  

%%%========= New Section =========
\section{EXPERIMENTAL EVALUATION} 
\label{sec:experiments}

This section outlines the evaluation protocols, implementation details, and training procedures of the proposed approach. Both quantitative and qualitative retrieval results are presented to support the proposed contributions. Additionally, a runtime analysis is performed to assess the computational efficiency of PointNetPGAP.

\subsection{The HORTO-3DLM Dataset}
\label{sec:dataset}

For real-world deployment, it is important to capture specific conditions that may induce unique challenges in agricultural environments. For instance, complex conditions may arise due to varying field layouts (row length, row width, open field, within tunnels, etc.), which can differ from plant type to plant type and even within the same plant species. Another aspect that may present challenges is the changing appearance of the same plantation due to plant growth and flowering stages.

Hence, we created a dataset called Horticulture 3D Localization and Mapping (HORTO-3DLM)\footnote{\href{https://github.com/Cybonic/HORTO-3DLM.git}{https://github.com/Cybonic/HORTO-3DLM.git}}, which was recorded in the United Kingdom over two years at different locations. The dataset includes data collected from four crops: two orchards (apples and cherries), and one strawberry plantation within polytunnels with a table-top growing system. All sequences incorporate various revisits.

The data collection process involved teleoperating a Clearpath Husky mobile robot equipped with a Velodyne VLP32 3D LiDAR (10Hz) and a ZED-F9P RTK-GPS (5Hz). Figure~\ref{fig:fronte} showcases the geolocation, the 3D maps of the four sequences, and the recording setup mounted on the mobile robot, while Table~\ref{tab:sequence} provides more detailed information regarding each sequence. The three sequences from orchards are denominated OJ22, ON22, and OJ23, while the sequence from the strawberry plantation is designated SJ23.

\begin{table}[t]
  \centering
  \caption{Summary of the four sequences. The columns have the following meaning: name of the sequence (Seq), month of recording (M), year of recording (Y), number of rows each sequence comprises (N.Rs), number of segments each sequence comprises (N.Segs), traveled distance (D) and plantation type.}
  {\renewcommand{\arraystretch}{1}% for the vertical padding
        \begin{tabular}{
        >{\centering\arraybackslash}p{0.08\columnwidth}
        >{\centering\arraybackslash}p{0.05\columnwidth}
        >{\centering\arraybackslash}p{0.05\columnwidth}
        >{\centering\arraybackslash}p{0.08\columnwidth}
        >{\centering\arraybackslash}p{0.05\columnwidth}
        >{\centering\arraybackslash}p{0.05\columnwidth}
        >{\centering\arraybackslash}p{0.05\columnwidth}
        >{\centering\arraybackslash}p{0.2\columnwidth}}
        \noalign{\hrule height 1pt}\hline	
        \addlinespace[0.1cm]
        Seq. & M & Y & NºScans & N.Rs & N.Segs & D [m] & Plantation Type\\
        \midrule
		\midrule
        \textbf{Orchards} & & & &\\
        \addlinespace[0.1cm]
        ON22 & Nov. & 2022 & 7974 & 4 & 6& 514& Apple (open)\\
        OJ22 & July   & 2022 & 4361  & 3 &  5&206 & Apple (open)\\
        OJ23 & June   & 2023 & 7229  & 3 & 5 &459 & Cherry (open)\\
        \hline
        \addlinespace[0.1cm]
         \textbf{Strawberries}  & & & &\\
         \addlinespace[0.1cm]
        SJ23 & June   & 2023 & 6389 & 3 & 5& 742 & tunnels \\
       \noalign{\hrule height 1pt}\hline	
        \end{tabular}%
    % \end{adjustbox}
    }
  \label{tab:sequence}%
\end{table}%

\begin{table}[t]
  \centering
  \caption{Cross-validation training and evaluation data. Each column represents the training and test data of the evaluated sequence (row 1)}
        {\renewcommand{\arraystretch}{1}% for the vertical padding
	%\begin{adjustbox}{max width=\columnwidth}
        \begin{tabular}{l|cccc}
        \noalign{\hrule height 1pt}\hline
                    &  OJ22 & OJ23  & ON22 &  SJ23 \\ \midrule
        Nº Training  Anchors ($r_{th} \leq 2m$)  & 1605  & 4801  & 1210 &  1107\\
        Nº Test Anchors  ($r_{th} \leq 10\text{m}$)     & 1797  & 1193 & 4993  & 3509\\
        %SJ23   & 3509  & 1107  & 598 \\
        \noalign{\hrule height 1pt}\hline	 
    \end{tabular}%
    }
    \label{tab:experiments}%
\end{table}%

\begin{table}[t]
  \centering
  \caption{ KITTI Odometry dataset and cross-domain training and evaluation data}
	\begin{adjustbox}{max width=\columnwidth}
        \begin{tabular}{l|ccccc}
        \noalign{\hrule height 1pt}\toprule

        \multicolumn{6}{l}{\textbf{KITTI Ondemtry dataset}}  \\  \midrule
        & 00     & 02     & 05     & 06     & 08 \\
        \midrule
		\midrule
         Nºof Frames &  4051 &  4661 & 2761 &   1101  & 4071 \\
         Nº of Anchors ($r_{th} \leq 10\text{m}$) &  809 &  344  & 454  &   273   &  353 \\ \bottomrule \bottomrule   \addlinespace[0.05cm]
         \multicolumn{6}{l}{\textbf{Cross Domain Training / Evaluation data}} \\  \midrule
                     & \multicolumn{3}{c}{Sequences} & \multicolumn{2}{c}{Nº Anchors}\\ \midrule \midrule
         Training    & \multicolumn{3}{c}{OJ22, OJ23, SJ23 ON22}  &\multicolumn{2}{c}{1705}\\
         Validation  & \multicolumn{3}{c}{00} &\multicolumn{2}{c}{809}\\
         Test        & \multicolumn{3}{c}{02, 05, 06, 08}&\multicolumn{2}{c}{1424}\\
          \noalign{\hrule height 1pt}\toprule	
        \end{tabular}%
    \end{adjustbox}
   % }
  \label{tab:kitti}%
\end{table}%

\subsubsection{HORTO-3DLM Dataset for Place Recognition}
\label{sec:gt}
From the proposed HORTO-3DLM dataset described in Section~\ref{sec:dataset}, each sequence was split into various segments (S1, S2, ..., S6) as illustrated in Fig.~\ref{fig:rows}. Besides the rows, we also included the extremities of the fields (\ie, S3 and S4) as segments. Thus, for each data point (LiDAR scan), there exists a corresponding segment class.

Following the protocol defined in Section~\ref{sec:b1}, $r_{th}$ was set to 2\,m for the training data, which means that a positive pair exists whenever an anchor-positive pair is within a range of 2\,m. Additionally, from all training anchor-positive pairs, only the anchors that are at least 0.5\,m apart were selected for training. 

\subsection{Evaluation Protocols}
The proposed approach is evaluated based on two strategies: a 4-fold cross-validation protocol; and a cross-domain approach.
Performance metrics are reported using Recall for the top-k retrieved candidates (\ie, Recall@k). 

\subsubsection{Cross-Validation} \label{sec:crossvalidation}
This protocol evaluates generalization capabilities within the same domain. The HORTO-3DLM dataset is split into four sequences, with models trained on three sequences and evaluated on the remaining one. This process is repeated four times to cover all sequences. The dataset split is outlined in Table \ref{tab:experiments}. A retrieved candidate is considered a true positive if it is within an $10$\,m radius of the query and belongs to the same segment; otherwise, it is considered a false positive.

\subsubsection{Cross-Domain}
This protocol assesses generalization capabilities across domains, specifically from horticultural to urban environments. The models are trained on the HORTO-3DLM dataset and evaluated on the KITTI Odometry dataset~\cite{Geiger2012CVPR}. The data split regarding this evaluation strategy is presented in Table \ref{tab:kitti}. A retrieved candidate is considered a true positive if it is within an $10$\,m radius of the query (no segments are considered); otherwise, it is considered a false positive.

\subsection{Training and Implementation Details}\label{sec:implementation}
All models were trained and evaluated under the same conditions. The input scans were downsampled to 10k points, and randomly rotated around the z-axis. The parameters of PointNetPGAP were the following: we used the original architecture of PoinNet, with T-Net disabled, descriptor size was set to 256 dimensions ($d=256$), the local features were set to 16 dimensions ($c=16$), the MLP was configured with three layers with 256, 64, and 6 neurons, respectively, while the weight of the final loss was set to 0.5 ($\alpha  = 0.5$). The SOTA models were implemented based on the original code.

All models were trained with 200 epochs, using early stopping to obtain the best performance on the validation set. All models were trained on an NVIDIA GeForce RTX 3090 GPU, using the closest positive and 20 negatives ($m=20$) for each anchor.  The margin value of the contrastive loss was set to 0.5 ($m=0.5$), and the model parameters were optimized using the AdamW optimizer with a learning rate ($L_r$) of 0.0001 and a weight decay ($W_d$) of 0.0005. Moreover, all proposed experiments were conducted on Python 3.8, and PyTorch with CUDA 11.6.

\begin{table*}[th]
    \centering
    \caption{Recall@1 performance of the cross-validation protocol (same domain). Reporting the results of the models trained: (without SLC)/(with SLC). Bolded scores indicate the highest value between models trained without and with SLC, while underlined scores represent the highest score in each column (\ie, the best-performing model for each sequence).}
{\renewcommand{\arraystretch}{1.0}% for the vertical padding
\begin{adjustbox}{max width=\linewidth}
\begin{tabular}{l|cccc|c}
\noalign{\hrule height 1pt}\hline \addlinespace[0.05cm]
Model &OJ22&OJ23&ON22&SJ23&MEAN\\
\hline \hline \addlinespace[0.05cm]
%PointNetGAPv1&0.661&0.599&0.493&0.409\\
OverlapTransformer \cite{9785497}
&\textbf{0.084}/0.075$^{\color{red}-0.009}$
&0.153/\textbf{0.164}$^{\color{green}+0.011}$
&\textbf{0.221}/0.194$^{\color{red}-0.027}$
&\textbf{0.084}/0.073$^{\color{red}-0.011}$
&\textbf{0.136}/0.127$^{\color{red}-0.009}$
\\

LOGG3D \cite{9811753}
&0.357/\textbf{0.394}$^{\color{green}+0.037}$
&0.309/\textbf{0.329}$^{\color{green}+0.020}$
&0.412/\textbf{0.470}$^{\color{green}+0.058}$
&0.254/\textbf{0.285}$^{\color{green}+0.031}$
&0.333/\textbf{0.370}$^{\color{green}+0.037}$
\\

PointNetVLAD \cite{angelina2018pointnetvlad}
&0.560/\textbf{0.659}$^{\color{green}+0.099}$
&0.485/\textbf{0.502}$^{\color{green}+0.017}$
&0.611/\textbf{0.625}$^{\color{green}+0.014}$
&0.676/\underline{\textbf{0.717}}$^{\color{green}+0.041}$
&0.583/\textbf{0.630}$^{\color{green}+0.043}$
\\

PNPGAP(ours) 
&\underline{0.873}/\underline{\textbf{0.880}}$^{\color{green}+0.007}$
&\underline{0.643}/\underline{\textbf{0.688}}$^{\color{green}+0.045}$
&\underline{0.685}/\underline{\textbf{0.734}}$^{\color{green}+0.049}$
&\underline{0.678}/\textbf{0.712}$^{\color{green}+0.034}$
&\underline{0.719}/\underline{\textbf{0.754}}$^{\color{green}+0.034}$
\\
\noalign{\hrule height 1pt}\hline

\end{tabular}
\end{adjustbox}
}
\label{tab:comparison}
\end{table*}

\subsection{Retrieval Evaluation and Discussion}

This section presents and discusses the empirical results of the work. First, an ablation study is conducted on the proposed PointNetPGAP model to showcase the contribution of each aggregator to the overall performance. Subsequently, PointNetPGAP is benchmarked within the same domain, both with and without the SLC model, against SOTA models: PointNetVLAD \cite{angelina2018pointnetvlad}, LOGG3D-Net \cite{9811753}, and OverlapTransformer \cite{9785497}. These models employ different strategies for processing LiDAR scans: PointNetVLAD processes scans directly, OverlapTransformer processes scans as range images, and LOGG3D-Net utilizes a voxel-based representation. Finally, the generalization capabilities of the proposed approach across domains are evaluated.

Throughout this section, PointNetPGAP is abbreviated as PNPGAP, and PNPGAP-SLC when trained with the SLC model.

\begin{table}[t]
    \centering
    \caption{Ablation study of PNPGAP, where GAP represents the global average pooling aggregator and PFI represents the aggregator based on pairwise feature interactions. The bolded scores represent the highest values in each column.}
{\renewcommand{\arraystretch}{1.3}% for the vertical padding
\begin{adjustbox}{max width=\linewidth}
\begin{tabular}{
>{\centering\arraybackslash}p{0.03\columnwidth}
>{\centering\arraybackslash}p{0.03\columnwidth}|
>{\centering\arraybackslash}p{0.04\columnwidth}
>{\centering\arraybackslash}p{0.04\columnwidth}
>{\centering\arraybackslash}p{0.04\columnwidth}
>{\centering\arraybackslash}p{0.05\columnwidth}
>{\centering\arraybackslash}p{0.07\columnwidth}|
>{\centering\arraybackslash}p{0.04\columnwidth}
>{\centering\arraybackslash}p{0.04\columnwidth}
>{\centering\arraybackslash}p{0.04\columnwidth}
>{\centering\arraybackslash}p{0.05\columnwidth}
>{\centering\arraybackslash}p{0.07\columnwidth}}
\noalign{\hrule height 1pt}\hline
\multirow{2}{*}{GAP} & \multirow{2}{*}{PFI} & \multicolumn{5}{c|}{Recall@1}  &  \multicolumn{5}{c}{Recall@1\%}\\
                &                & OJ22 & OJ23 & ON22 & SJ24 & MEAN & OJ22 & OJ23 & ON22 & SJ24 &MEAN \\\hline\hline
 \checkmark    &                 & 0.819& 0.589&0.687 &0.714 & \cellcolor[gray]{0.85} 0.701& 0.928& 0.909& 0.962& 0.975&\cellcolor[gray]{0.85} 0.944      \\
               &  \checkmark     & 0.827& 0.613&0.685 &0.713 & \cellcolor[gray]{0.85} 0.710& \textbf{0.962}& 0.925& 0.950& \textbf{0.988}& \cellcolor[gray]{0.85} 0.956\\
\checkmark     &  \checkmark     & \textbf{0.880}& \textbf{0.688}&\textbf{0.734} &\textbf{0.723}&\cellcolor[gray]{0.85}\textbf{0.756} & 0.943& \textbf{0.938} & \textbf{0.985}& 0.983 & \cellcolor[gray]{0.85} \textbf{0.962}\\
\noalign{\hrule height 1pt}\hline
\end{tabular}
\end{adjustbox}
}
\label{tab:ablation}
\end{table}

\subsubsection{Ablation Study}

The ablation study is conducted following the cross-validation protocol outlined in Section~\ref{sec:crossvalidation}, with results presented in Table~\ref{tab:ablation}. For this analysis, we report Recall@1 and Recall@1\% for each aggregation approach, both individually and combined. The results suggest that combining both aggregators results in superior performance compared to using each aggregator separately, with an improvement of 0.6 percentage points (pp) in retrieving the Top-1\% candidates and 4.7 pp for retrieving the Top-1 candidate. These findings indicate that the proposed aggregator significantly enhances the model's ability to identify the nearest neighbor.

\begin{figure}[t]
 \centering
    \begin{subfigure}[b]{0.49\columnwidth}
    OJ22
    \centering
    \includegraphics[width=\columnwidth, trim={\leftvertical cm \bottom cm \rightvertical cm \top cm},clip]{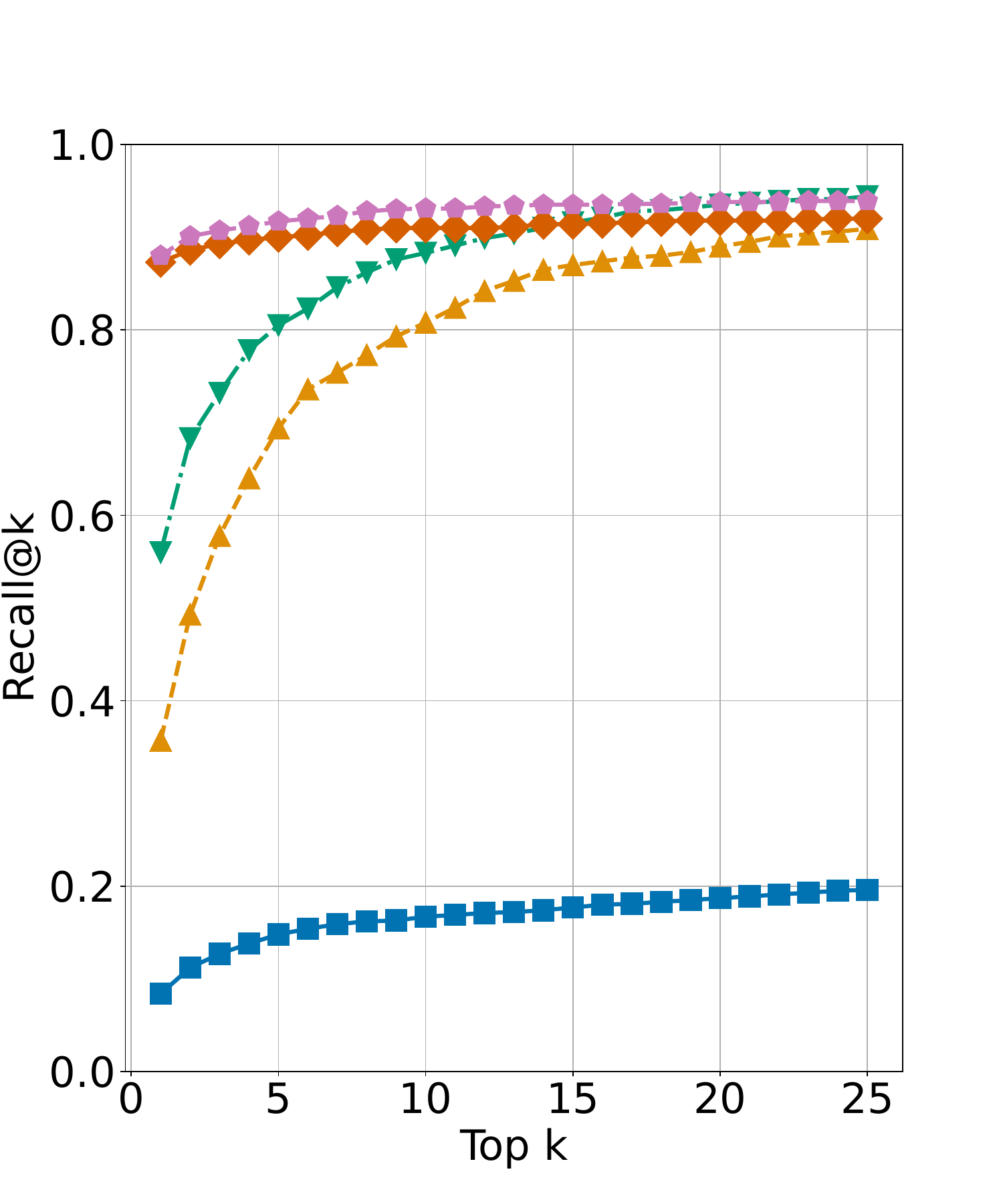}
    %\caption{}
  \end{subfigure}
  \hfill
  \begin{subfigure}[b]{0.49\columnwidth}
    OJ23
    \centering
    \includegraphics[width=\columnwidth, trim={\leftvertical cm \bottom cm \rightvertical cm \top cm},clip]{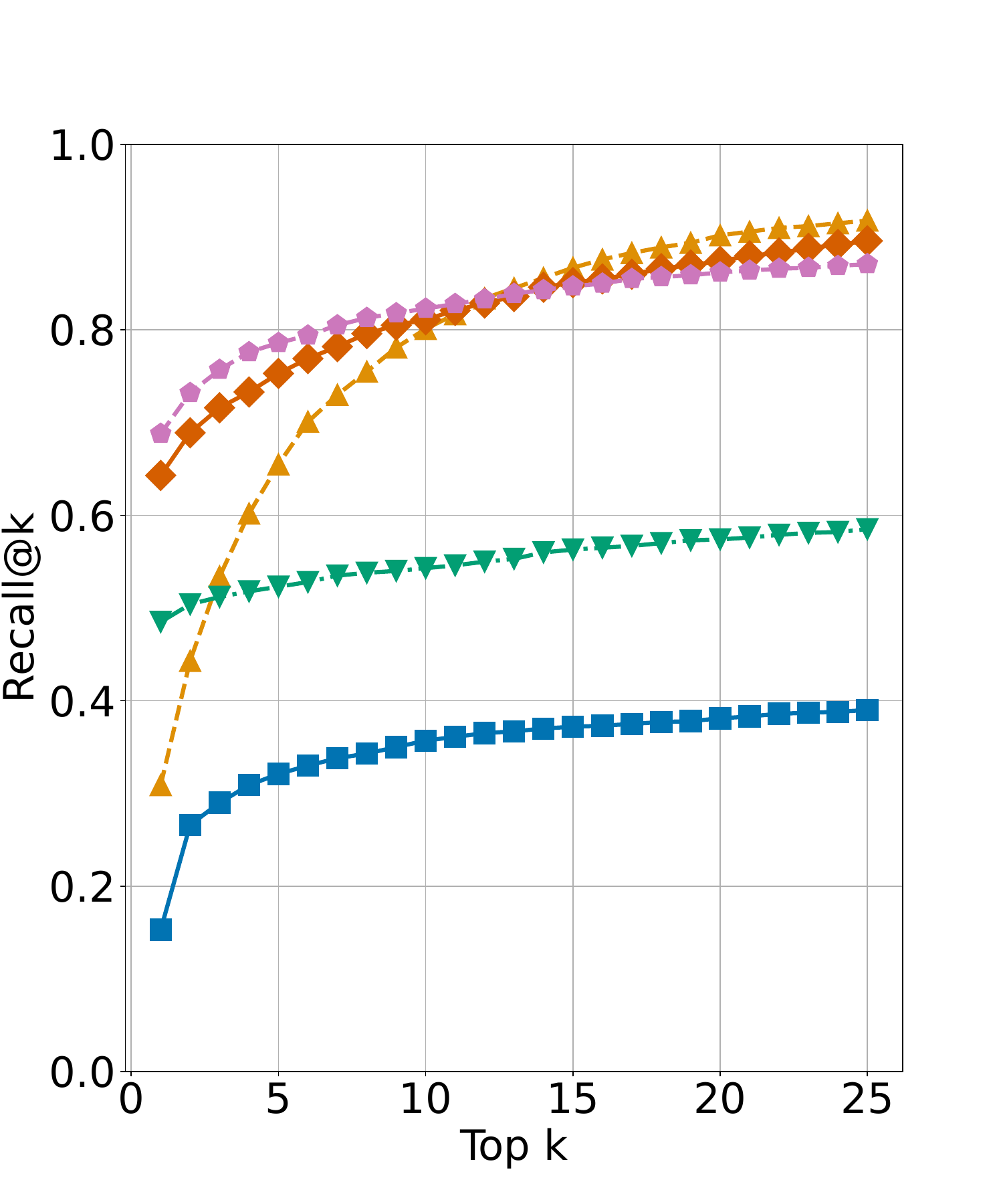}
    %\caption{}
  \end{subfigure}
  \hfill
  \begin{subfigure}[b]{0.49\columnwidth}
    ON22
    \centering
    \includegraphics[width=\columnwidth, trim={\leftvertical cm \bottom cm \rightvertical cm \top cm},clip]{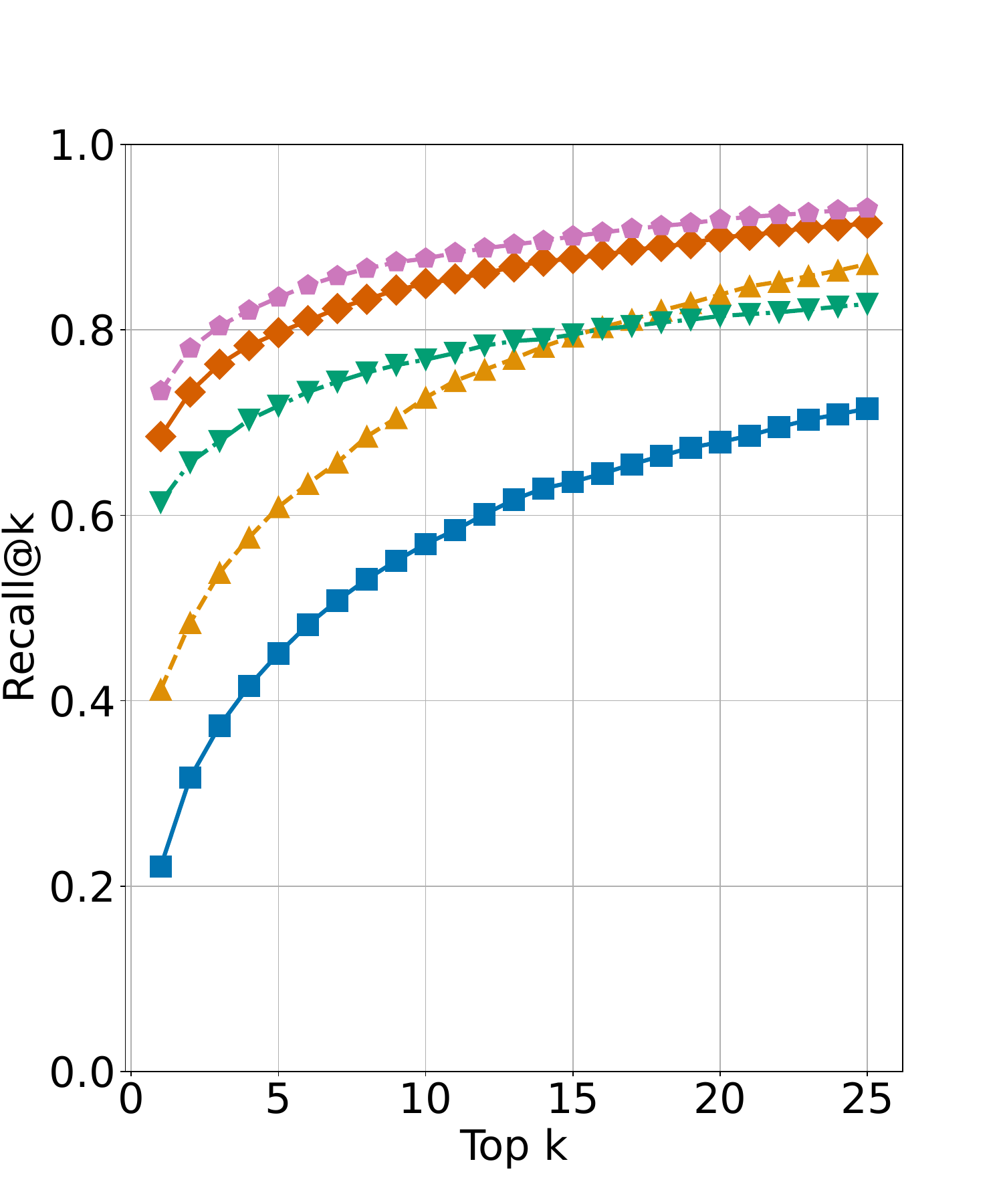}
    %\caption{}
  \end{subfigure}
  \hfill
  \begin{subfigure}[b]{0.49\columnwidth}
    SJ23
    \centering
    \includegraphics[width=\textwidth, trim={\leftvertical cm \bottom cm \rightvertical cm \top cm},clip]{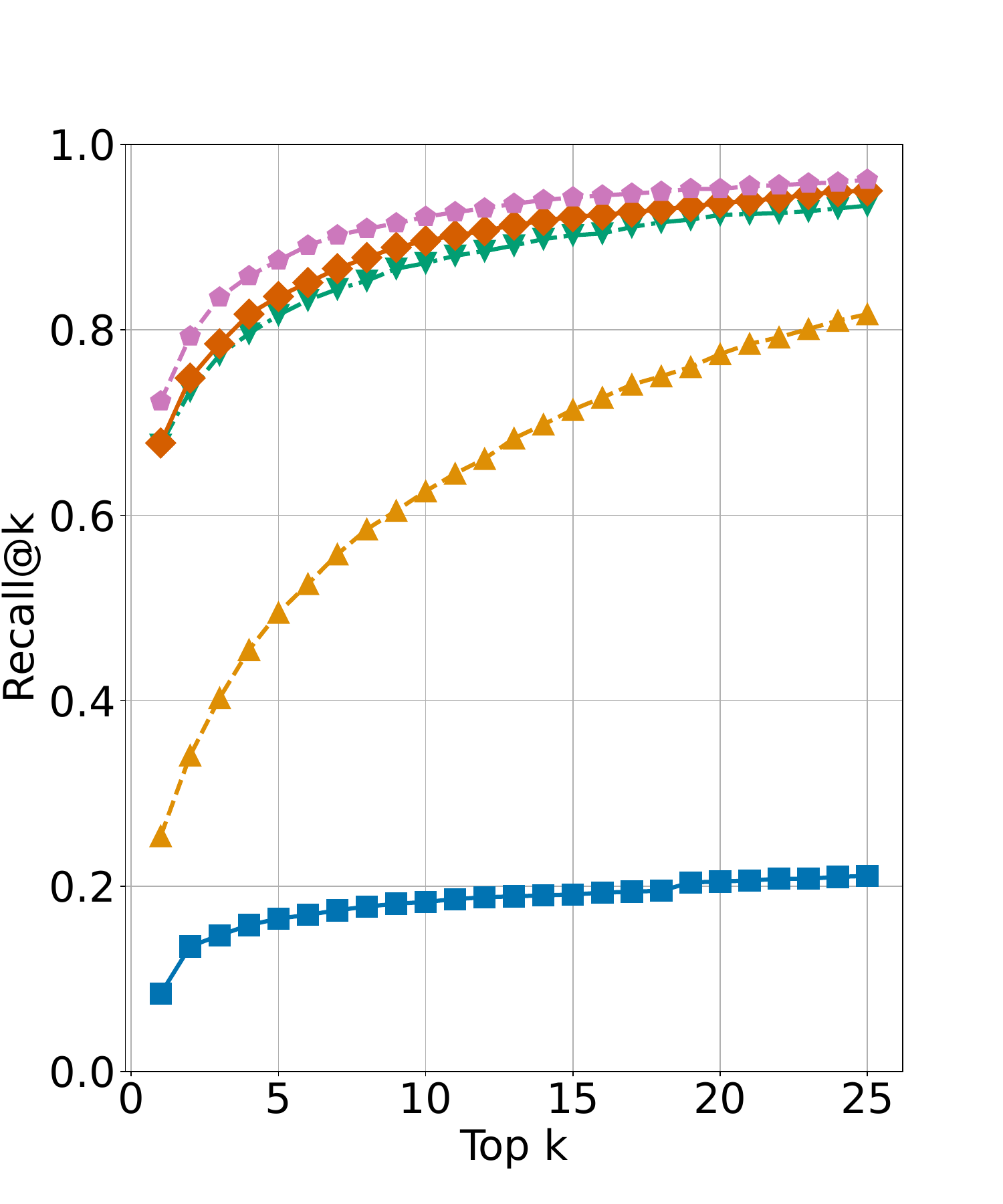}
    %\caption{}
  \end{subfigure}
   \hfill
  \begin{subfigure}[b]{1.\columnwidth}
    \centering
    \includegraphics[width=\columnwidth, trim={0cm 0cm 0cm 0cm},clip]{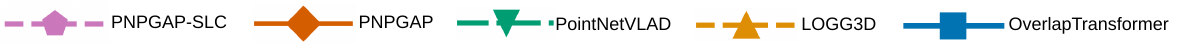}
    %\caption{}
  \end{subfigure}
  %}
  \caption{Retrieval performance for the top-25 candidates on the four sequences.}
  \label{fig:results}
\end{figure}

\begin{figure*}[!ht]
 \centering
\includegraphics[width=\textwidth, trim={0.0cm 1cm 0cm 1.5cm},clip]{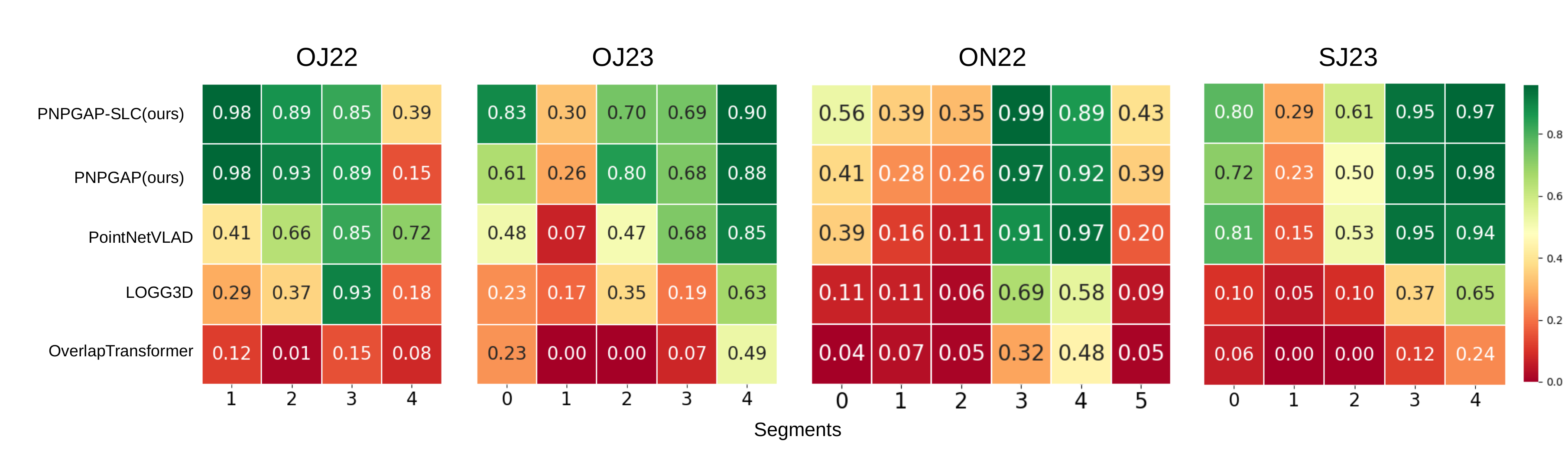}
\caption{Recall@1 performance at the segment-level for the four sequences. For more details on the segments, please see Fig.~\ref{fig:rows}.}
\label{fig:segments}
\end{figure*}

\subsubsection{Cross-Validation (same domain)} 
\begin{figure}[t]
    \centering
    \includegraphics[width=\linewidth, trim={0cm 0cm 9cm 0cm},clip]{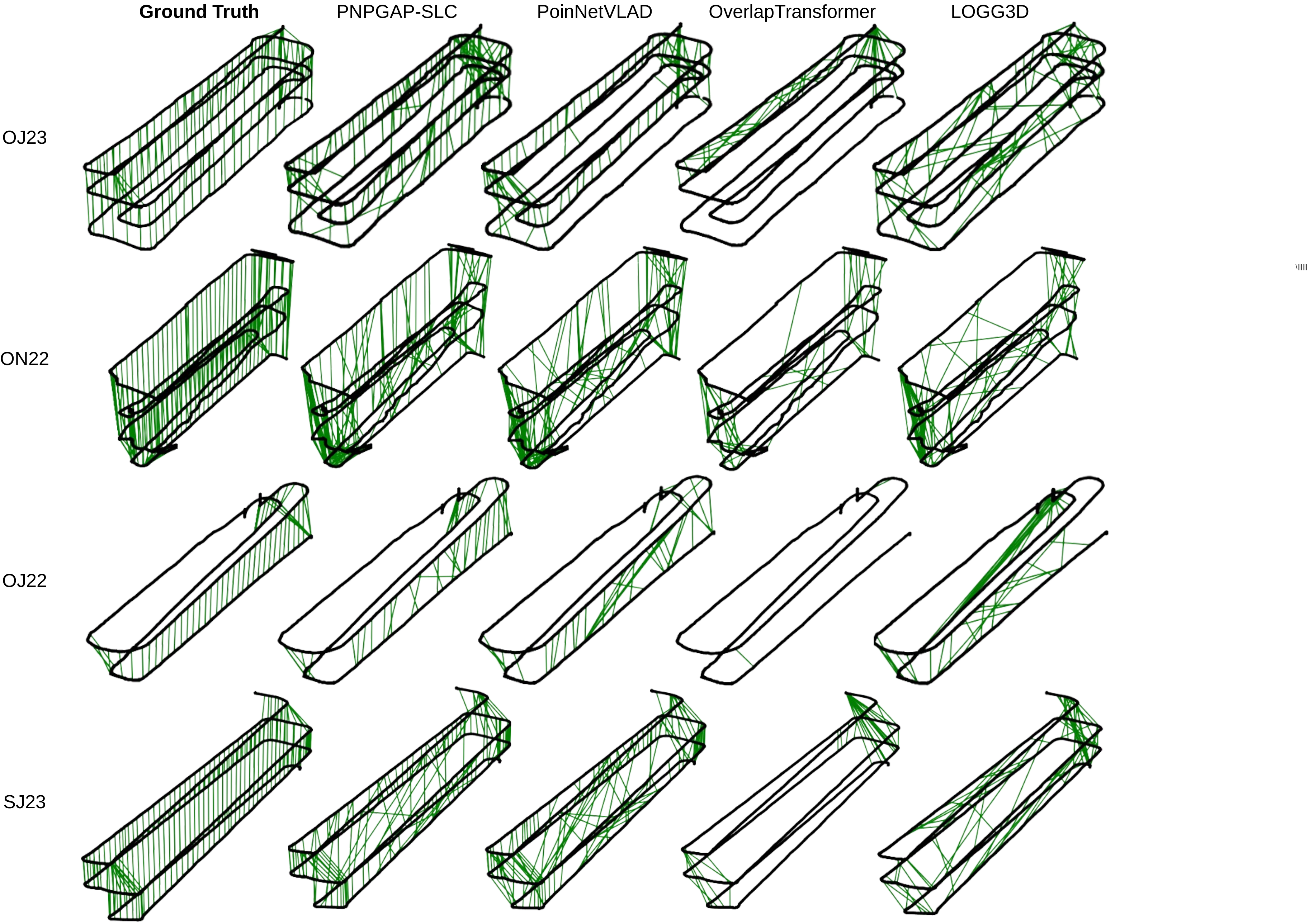}
    \caption{True positive predictions of the top-1 retrieved candidates.}
    \label{fig:qualitative}
\end{figure}

The results of this experiment are presented in Table~\ref{tab:comparison}. PNPGAP emerges as the best model when compared to the SOTA models, outperforming PointNetVLAD by 13.6 pp without the SLC model and by 12.4 pp with the SLC model when retrieving the Top-1 candidate.This superiority is illustrated further in Fig.~\ref{fig:qualitative}, where a qualitative assessment shows the true positive predictions of each model for the Top-1 candidate retrieval along the paths. Given that PNPGAP and PointNetVLAD share the same backbone, these results suggest that the proposed aggregator is more effective in horticultural environments, particularly in handling sparse scans, than other first- and second-order aggregators.

When analyzing the performance of retrieving the Top-25 candidates (Fig.~\ref{fig:results}), LOGG3D-Net, which relies on a voxel-based network for feature extraction, exhibits the highest increase in performance. Meanwhile, OverlapTransformer, which relies on range images and on a CNN-based network for feature extraction, shows poor performance across all sequences. The low performance of OverlapTransformer is also apparent in Fig.~\ref{fig:qualitative}, where in some segments the model fails to predict any loops correctly. This suggests the unsuitability of this model for agricultural environments.

When adding the SLC model at training time (as shown in Table~\ref{tab:comparison}), the results indicate that this additional training signal enhances performance across all sequences and models (except for OverlapTransformer), with an average performance increase of 3.7 pp for LOGG3D, 4.3 pp for PointNetVLAD, and 3.34 pp for the proposed PNPGAP model. 

At the segment-level (see Fig.~\ref{fig:segments}), most models (without SLC) achieve their highest performance in segments 3 and 4, which represent the extremities of the crops. As illustrated in Fig.~\ref{fig:fronte},these segments are more distinctive. In contrast, segments representing the rows exhibit greater similarity, creating ambiguities among adjacent rows, which leads to lower retrieval performance.

However, when the SLC model is incorporated, the results at the segment level indicate a notable performance increase, particularly in the segments corresponding to rows. This improvement suggests that the SLC model produces more robust descriptors that are less susceptible to ambiguities.

\begin{table*}[t]
    \centering
    \caption{Recall@1 of the cross-domain evaluation: (without SLC)/(with SLC). The bolded scores represent the highest value between trained without and with SLC, while the underlined scores indicate the highest score in each column.}
    \label{tab:crossdomain}

\begin{adjustbox}{max width=\linewidth}
\begin{tabular}{l|c|ccccc}
\noalign{\hrule height 1pt}\hline
\addlinespace[0.01cm]
& Validation & \multicolumn{5}{c}{Testing} \\  \addlinespace[0.01cm] \cline{2-7} \addlinespace[0.05cm] % 
\multirow{2}{*}{} &00&02&05&06&08&MEAN\\
\hline \hline \addlinespace[0.05cm] 
Overlap Transformer \cite{9785497} &
0.247/\textbf{0.278}$^{\color{green}+0.031}$&
0.192/\textbf{0.253}$^{\color{green}+0.061}$&
0.264/\textbf{0.361}$^{\color{green}+0.097}$&
0.227/\textbf{0.355}$^{\color{green}+0.128}$&
0.023/\textbf{0.040}$^{\color{green}+0.017}$&
0.176/\textbf{0.252}$^{\color{green}+0.076}$\\

%LOGG3D&0.043&0.084&0.108&0.271&0.043&0.110&0.084\\
%LOGG3DLoss&0.083&0.116&0.176&0.209&0.043&0.125&0.060\\
LOGG3D \cite{9811753}&
0.043/\textbf{0.083}$^{\color{green}+0.040}$&
0.084/\textbf{0.116}$^{\color{green}+0.032}$&
0.108/\textbf{0.176}$^{\color{green}+0.068}$&
\textbf{0.271}/0.209$^{\color{red}-0.062}$&
\textbf{0.043}/\textbf{0.043}$^{\color{green}+0.000}$&
0.127/\textbf{0.138}$^{\color{green}+0.010}$\\

PointNetVLAD  \cite{angelina2018pointnetvlad}
&\underline{0.841}/\underline{\textbf{0.845}}$^{\color{green}+0.004}$&
\underline{0.442}/\underline{\textbf{0.451}}$^{\color{green}+0.009}$&
0.663/\underline{\textbf{0.738}}$^{\color{green}+0.075}$&
0.934/\underline{\textbf{0.978}}$^{\color{green}+0.044}$&
0.123/\textbf{0.234}$^{\color{green}+0.111}$&
0.541/\underline{\textbf{0.600}}$^{\color{green}+0.060}$\\

PNPGAP(ours)&
0.826/\textbf{0.841}$^{\color{green}+0.015}$&
0.395/\textbf{0.448}$^{\color{green}+0.053}$&
\underline{\textbf{0.683}}/0.659$^{\color{red}-0.024}$&
\underline{0.938}/\textbf{0.971}$^{\color{green}+0.033}$&
\underline{0.248}/\underline{\textbf{0.308}}$^{\color{green}+0.060}$&
\underline{0.566}/\textbf{0.597}$^{\color{green}+0.031}$\\

\noalign{\hrule height 1pt}\hline
\end{tabular}
\end{adjustbox}
\end{table*}

\subsubsection{Cross-Domain} \label{sec:cross_domain}

The results of the cross-domain experiment are presented in Table~\ref{tab:crossdomain}. 
In this experiment, the performance difference between PNPGAP and PointNetVLAD is smaller, with PNPGAP outperforming by 3.3~pp. 
This difference is primarily attributable to PNPGAP’s superior performance on sequence 08, the most challenging sequence, where most revisits occur from the reverse direction.
This finding suggests that, while the proposed aggregation approach provides less of an advantage in urban environments compared to horticultural settings, it still demonstrates superior rotation-invariant properties relative to other SOTA models, particularly PointNetVLAD.

When incorporating the SLC model during training, the performance improvements observed across models are consistent with those seen within the same domain, with the exception of OverlapTransformer. Notably, OverlapTransformer’s performance is significantly higher in the cross-domain experiment compared to its results on the HORTO-3DLM dataset, despite being trained on the same dataset in both cases. We hypothesize that this performance gain is primarily due to OverlapTransformer’s input representation, which appears to be better suited to the structured nature of urban environments.

%\vspace{-0.4cm}
\subsection{Runtime Analysis}\label{sec:runtime}
Processing capacity and model size are critical factors for real-world implementation, particularly in resource-constrained robotics applications. The runtime experiment results, presented in Table~\ref{tab:runtime}, are obtained using the hardware detailed in Section~\ref{sec:implementation}. The reported times, measured in milliseconds (ms), represent the duration each model requires to process a batch of 20 scans. Retrieval duration was excluded from this experiment, as it remains consistent across all models. Model size is expressed by the number of parameters, in millions (M). PNPGAP is notably compact, with 0.4 million parameters, and processes a batch in an average of 4.4 ms.

Comparative analysis with other SOTA models reveals that PNPGAP is approximately 5\% the size of LOGG3D-Net and 31 times faster. Compared to PointNetVLAD, PNPGAP is only 2\% of its size and 6 times faster. These results underscore the computational efficiency of the proposed PNPGAP model.

\begin{table}[t]
  \centering
  \caption{The average runtime per batch, where the batch size is 20 scans. }
  {\renewcommand{\arraystretch}{1}% for the vertical padding
	%\begin{adjustbox}{max width=\columnwidth}
        \begin{tabular}{>{\centering\arraybackslash}p{0.25\columnwidth} |
                        >{\centering\arraybackslash}p{0.1\columnwidth} |
                        >{\centering\arraybackslash}p{0.06\columnwidth}
                        >{\centering\arraybackslash}p{0.06\columnwidth}
                        >{\centering\arraybackslash}p{0.06\columnwidth}
                        >{\centering\arraybackslash}p{0.06\columnwidth}
                        >{\centering\arraybackslash}p{0.08\columnwidth}}
 \noalign{\hrule height 1pt}\hline
                        
 & Param. [M]&	OJ22 [ms]&	OJ23 [ms]& ON22 [ms]& SJ23 [ms] &  MEAN [ms] \\ \hline \hline

OverlapTransformer &48.2&12.5&11.5&11.4&11.5&\cellcolor[gray]{0.9}{11.7}\\
LOGG3D&8.8&108.3&115.9&108.0&82.0&\cellcolor[gray]{0.95}{103.5}\\
PointNetVLAD	&19.8&	23.3&	22.1&	22.1&	22.4&	\cellcolor[gray]{0.85}{22.5}\\
PNPGAP\,(ours)	&0.4&	3.3&	3.5&	3.3&	3.4&	\cellcolor[gray]{0.68}{3.38}\\
\noalign{\hrule height 1pt}\hline
\end{tabular}%
}
\label{tab:runtime}
\end{table}%

%\vspace{-0.3cm}
\section{CONCLUSIONS} \label{sec:conclusion}
In this letter, we addressed the challenge of 3D LiDAR-based place recognition in horticultural environments by introducing the PointNetPGAP model, which integrates two statistically-inspired aggregators, along with an SLC model designed to mitigate inter-row descriptor ambiguities. We also presented the HORTO-3DLM dataset, comprising sequences from orchards and a strawberry plantation.

Experimental results obtained from the HORTO-3DLM dataset, using a cross-validation protocol, as well as from the KITTI Odometry dataset, suggest that PointNetPGAP outperforms SOTA models, as evidenced by an ablation study and benchmarks at both the sequence and segment levels. Furthermore, incorporating the SLC model during training consistently improved performance, particularly in segments with high inter-row ambiguity, highlighting the critical role of context-specific training signals. % in environments with significant LiDAR scan overlap.

%\vspace{-0.2cm}
%%%========= New Section =========
\section*{ACKNOWLEDGMENTS}
 The authors would also like to thank Dr Charles Whitfield at NIAB East Malling for facilitating orchard data collection campaigns.

%\vspace{-0.5cm}
%%%%%%%%%%%%%%%%%%%%%%%%%%%%%%%%%%%%%%%%%%%%%%%%%%%%%%%%%%%%%%%%%%%%%%%%%%%%%%%%

\bibliographystyle{IEEEtran}

\bibliography{ref}

\end{document}